\documentclass{article}

\usepackage{arxiv}

\usepackage[utf8]{inputenc} 
\usepackage[T1]{fontenc}    
\usepackage{hyperref}       
\usepackage{url}            
\usepackage{booktabs}       
\usepackage{amsfonts}       
\usepackage{nicefrac}       
\usepackage{microtype}      
\usepackage{lipsum}         
\usepackage{graphicx}
\usepackage{natbib}
\usepackage{doi}

\usepackage{subcaption}
\usepackage{amsmath}
\usepackage{cleveref}       

\title{Deep Instance Segmentation and Visual Servoing to Play Jenga with a Cost-Effective Robotic System}
\date{}

\author{Luca Marchionna\thanks{Equal contribution.} , Giulio Pugliese$^{*}$, \href{https://orcid.org/0000-0002-6204-3845}{Mauro Martini$^{1}$}, \href{https://orcid.org/0000-0002-4445-9783}{Simone Angarano$^{1}$},\\ \href{https://orcid.org/0000-0003-4744-4349}{\textbf{Francesco Salvetti}$^{1,2}$}, \href{https://orcid.org/0000-0002-1921-0126}{\textbf{Marcello Chiaberge}$^{1}$}\\
    $^1$PIC4SeR PoliTo Interdepartmental Center for Service Robotics\\
    $^2$SmartData@PoliTo, Big Data and Data Science Laboratory\\
    Department of Electronics and Telecommunications, Politecnico di Torino, Turin, Italy\\
	\texttt{\{lucamarchionna97,pugliese.giulio8\}@gmail.com,}\\
	\texttt{\{mauro.martini,simone.angarano,francesco.salvetti,marcello.chiaberge\}@polito.it}
}

\renewcommand{\headeright}{}
\renewcommand{\undertitle}{}
\renewcommand{\shorttitle}{Deep Instance Segmentation and Visual Servoing to Play Jenga with a Cost-Effective Robotic System}

\begin{document}
\maketitle

\begin{abstract}
    The game of Jenga represents an inspiring benchmark for developing innovative manipulation solutions for complex tasks. Indeed, it encouraged the study of novel robotics methods to successfully extract blocks from the tower. A Jenga game round undoubtedly embeds many traits of complex industrial or surgical manipulation tasks, requiring a multi-step strategy, the combination of visual and tactile data, and the highly precise motion of the robotic arm to perform a single block extraction. In this work, we propose a novel, cost-effective architecture for playing Jenga with e.Do, a 6-DOF anthropomorphic manipulator manufactured by Comau, a standard depth camera, and an inexpensive monodirectional force sensor. Our solution focuses on a visual-based control strategy to accurately align the end-effector with the desired block, enabling block extraction by pushing. To this aim, we train an instance segmentation deep learning model on a synthetic custom dataset to segment each piece of the Jenga tower, allowing visual tracking of the desired block's pose during the motion of the manipulator. We integrate the visual-based strategy with a 1D force sensor to detect whether the block can be safely removed by identifying a force threshold value. Our experimentation shows that our low-cost solution allows e.DO to precisely reach removable blocks and perform up to 14 consecutive extractions in a row.
\end{abstract}

\keywords{Jenga, Robotic Arm, Deep Instance Segmentation, Visual Servoing, Sensor Fusion} 


\section{Introduction}
\label{sec:introduction}
In recent years, visual-based control strategies have successfully spread in a wide variety of robotics contexts \cite{sun2018review}. Nowadays, advances in computer vision for robotic perception are strictly tied to deep learning (DL). DL has been used in many robotics applications where objects must be detected \cite{zhao2019object} or segmented \cite{zhang2018fast} to address a manipulation task, demonstrating competitive advantages compared to classic image processing algorithms in terms of accuracy and robustness. For instance, relevant works have been proposed in recent years in the precision agriculture field to support autonomous navigation \cite{martini2022position, salvetti2022waypoint}, harvesting \cite{harvesting}, and spraying \cite{spraygrape}. Intelligent DL-based behaviors are also desired for visual-based robotic surgery to detect and segment instruments \cite{kletz2019identifying, hasan2019u}, and in many industrial robotic tasks \cite{chen2018industrial, domae2019recent}.
Nonetheless, to fill the gap between robot and human perception, multisensory approaches have recently been studied and evolved in novel robotic platforms, combining visual data with vocal interfaces \cite{juel2020smooth, eirale2022marvin}, or tactile sensors \cite{yu2020human, goldau2019autonomous, dong2022lifelong}. 

\begin{figure*}[t]
    \centering
    \includegraphics[width=\textwidth]{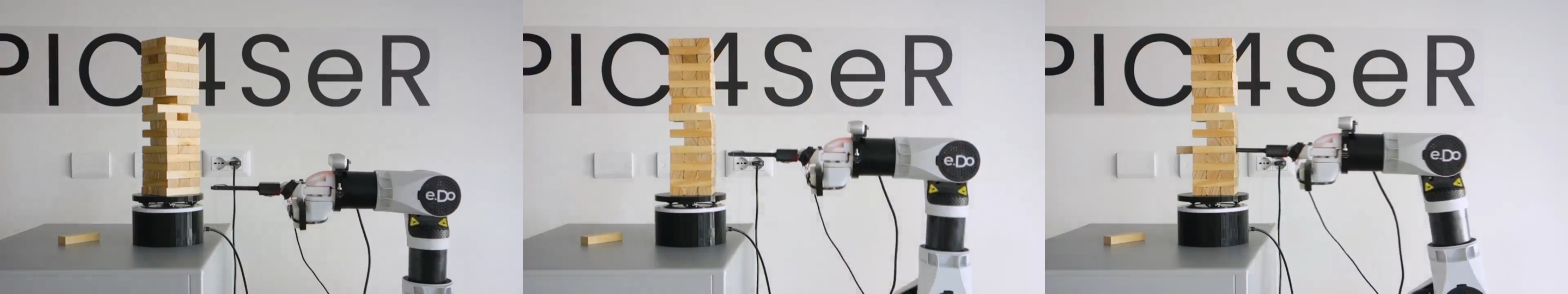}
    \caption{Our proposed solution in action: from left to right in the sequence, the e.DO robot selects a block to extract in the tower, adopting a visual-based control to approach it. Then, it starts pushing for the block extraction if it detects a low reaction force.}
    \label{fig:edo_frames}
\end{figure*}

The game of Jenga can be a perfect example of a challenging benchmark for robotic perception and control. In recent years, researchers have tackled the game with disparate platforms and approaches, adopting sophisticated manipulators \cite{manipulator_plays_jenga, multiarticulated_jenga} and complex control systems \cite{mit_jenga, deepQ_jenga}. The contribution to an effective robotic solution for Jenga goes beyond the fascinating dynamics of this popular game. Indeed, it can support the evolution of cutting-edge visual and multisensory control strategies for complex real-world tasks requiring human-level precision. The case of Jenga is not isolated in the historical advancement of Artificial Intelligence (AI), where games are often used as a common benchmark for newly proposed learning algorithms \cite{justesen2019deep,mnih2013playing,silver2017mastering}. The complexity of a round of Jenga resides in two different factors: first, it requires a multi-step policy to select a feasible block in the tower, approach it and finalize its extraction. Second, all these steps are based on the combination of real-time visual and tactile data processing and the highly precise motion of the end-effector for a single block extraction. According to this, it can be surely compared to real-world industrial \cite{domae2019recent}, surgical \cite{caccianiga2021multi}, or agricultural \cite{zheng2021mango, zheng2021dexterous} manipulation tasks. 

This work presents a cost-effective system to play Jenga with the educational robotic manipulator, e.DO by Comau and a custom pushing finger as an end-effector. Our proposed solution is based on the combination of visual and tactile perception to handle the human-level complexity and the high precision required by the task. In particular, compared to previous attempts to play Jenga with a manipulator, we adopted a single RGB-D camera and a basic 1-D force sensor as complementary hardware to the robot arm, considerably reducing the cost and complexity of the solution. An illustrative sequence of frames depicting our robotic system in action is shown in Figure \ref{fig:edo_frames}. Overall, our perception and control system is composed of the following:

\begin{itemize}
    \item a DL-based Instance Segmentation model fine-tuned on a custom Jenga tower dataset realized in simulation, which effectively allows the system to segment and select single blocks; 
    \item an eye-in-hand visual control strategy that carefully handles the blocks extraction;
    \item a 1-D force sensor to correctly evaluate the removability of a specific block.
\end{itemize}

Our extensive experimentation validates all the sub-components of the proposed system. First, we study the force reaction on the 1-D sensor and obtain a threshold-based decision policy to classify the extraction of the block as feasible or not. Then, we provide details on the training and testing of the Instance Segmentation model, comparing results obtained on simulated and real-world images. Moreover, we test our visual perception pipeline composed of segmentation and pose estimation of a group of blocks, measuring the tracking time during several runs. The adopted visual servoing control strategy is tested on the two major behavioral features of interest: precision, in terms of distance between the point of contact and the center of the block, and efficiency, estimated as the time required to align the end-effector to a block. Finally, we evaluate the overall performance of our solution by counting consecutive successful block extractions.

The paper is organized as follows. In Section \ref{sec:related}, related works are presented in three subsections discussing previous attempts to play Jenga with a manipulator and the state-of-the-art of deep instance segmentation, visual servoing, and multisensory control strategy in robotics applications. Section \ref{sec:methodology} illustrates the overall strategy adopted to tackle the Jenga game and all the specific components of our solution. Finally, in Section \ref{sec:experiments}, we present and discuss all the conducted experiments and the obtained results. Section \ref{sec:conclusion} draws some conclusions and potential suggestions for future works.

\begin{figure*}[t]
    \centering
    \includegraphics[width=\textwidth]{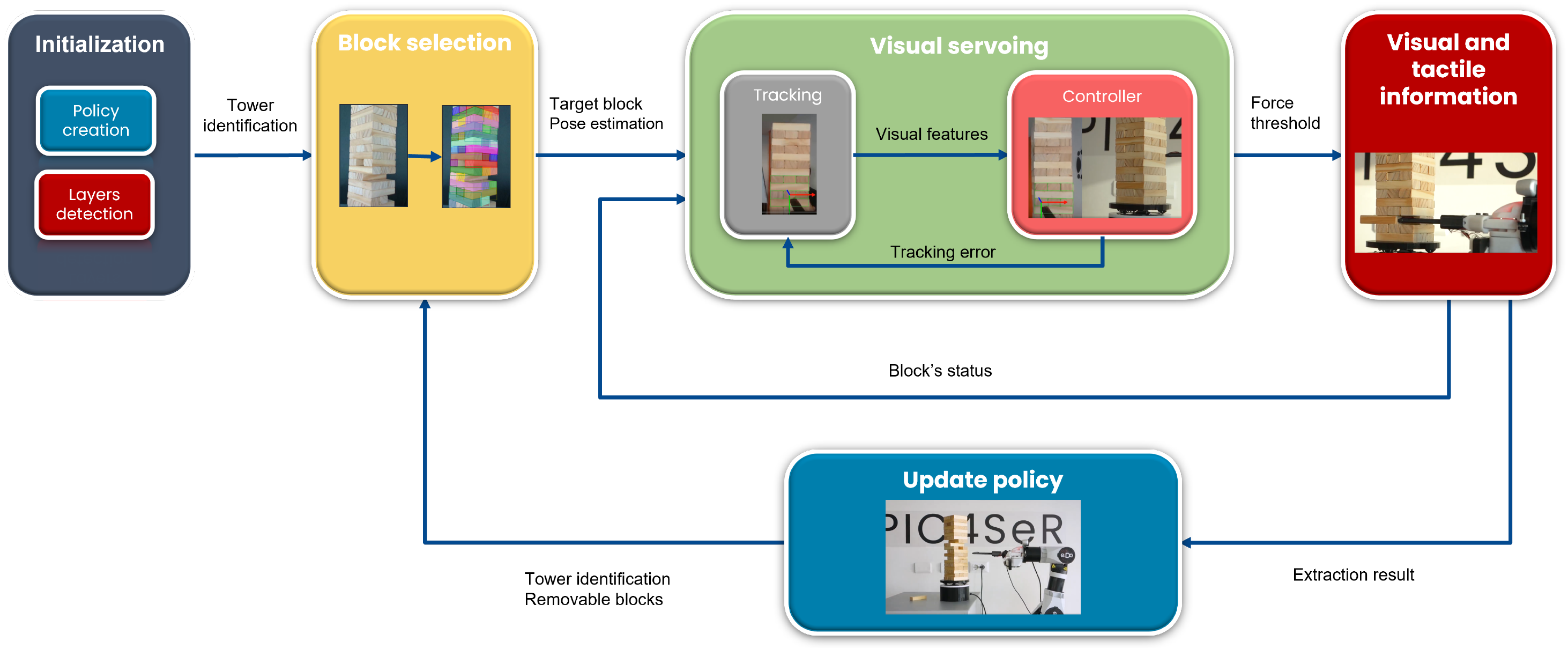}
    \caption{Block diagram of the proposed system architecture for Jenga: after tower identification, an RGB image of the tower is first used to extract the segmentation masks of each block with the Deep Instance Segmentation neural network. A block selection policy chooses the block to extract while considering blocks that have already been tried. The selected block is precisely approached by the end-effector adopting a visual-based control strategy. Finally, the end-effector pushes the block, and an inexpensive 1-D force sensor verifies its removability to stop or approve the extraction attempt.}
    \label{fig:full-scheme}
\end{figure*}

\section{Related Works}
\label{sec:related}
In this section, we first describe previous studies aimed at playing Jenga with a robot. Then we introduce the computer vision task of Instance Segmentation and its state-of-the-art and report similar works adopting visual servoing and multisensory control in robotic applications. 

\subsection{Playing Jenga with a robot}
\label{sec:related_jenga}
Jenga is a common benchmark for robotic systems, allowing for a direct comparison of methods, experiments, and results. So far, few works have proposed a complete visual-based robotic system to play the game autonomously. At the same time, several studies focus on a partial aspect of the game with a specific solution.

Recently and most notably, Fazeli et al. \cite{mit_jenga} delved into the details of the manipulation and artificial intelligence capabilities needed to learn Jenga by sight and touch, achieving 20 consecutive block extractions; their system is made up of an end-effector that can push and pick blocks, a long-reach 6-DOF industrial robot arm, an expensive 6-axis force sensor, and a fixed camera. The method is focused on learning from multisensory fusion and is compared to state-of-the-art learning paradigms with simulation and experiments, resulting in high performance and fast convergence. Their approach adheres to all the rules of Jenga with a tower in standard conditions and considers both block removal and placement on top of the tower. However, the hierarchical control strategy they adopt to carefully push the block during the extraction strongly relies on the use of an expensive 6-axis force sensor and a professional industrial arm, drastically increasing the overall cost of the solution.

Kroger et al.\cite{manipulator_plays_jenga}, who adopt similar expensive hardware, achieve 29 extractions with a 6-DOF industrial robot arm, a 6-axis force sensor, a 6-axis acceleration sensor, a laser distance sensor, and two static CCD cameras. The authors develop a modular control system based on a generic number of sensors with a primitive manipulation programming interface to play a standard Jenga game, manually recoloring the blocks to help the vision system.
Both reaction force and tower perturbed movements seen from cameras provide extraction feedback when a block is randomly chosen to be extracted. Pose estimation from cameras is refined with a laser distance profiler before gripping the block, which is then placed on top with force feedback. 

Wang et al.\cite{strategic_jenga} propose a simpler system to leverage inexpensive vision and manipulation hardware to develop a strategic planner based only on visual feedback. They achieve up to 5 consecutive extractions. The limitations of using classical computer vision with two CMOS cameras and a 5-DOF Pioneer short-reach robotic arm without force measurements led the authors to choose a quite different and simplified Jenga setup compared to the real one. Target blocks were partially pre-pulled and distanced one from the other, and the tower had half the levels.

A two-fingers, anthropomorphic, 7-DOF industrial robot arm is used in \cite{multiarticulated_jenga}. An eye-in-hand omnidirectional camera detects the tower configuration, and a block is chosen using a stability criterion. The block is grasped by the robot hand, which mounts a 6-axis force sensor on each fingertip. The Jenga setup is not standard, as the tower has only 10 layers. However, the system can detect and place blocks on the top of the tower, presenting a pretty high autonomy level.

A fine-grained kinematic analysis of the physics behind weight, friction interactions, and stability of the Jenga tower is studied in \cite{force_based_jenga}, both during and after the extraction. With a 6-DOF manipulator, a custom gripper, and a 6-axis force sensor, they compare the real forces with the modeled ones and achieve 14 consecutive extractions before breaking the tower. No vision or pose estimation system is used, so a human operator must provide poses and manually rotate the tower.

Differently, Yoshikawa et al.\cite{deepQ_jenga} investigated a Reinforcement Learning approach, using a deep Q-Network and a 6-DOF manipulator to correctly push a block without a priori knowledge of the kinematics and stability of the tower. The work is done in simulation on an ideal Jenga tower. Negative rewards are extrapolated from how humans play the game, such as pushing in the wrong direction and touching other blocks. The result is a policy for precise movement.

Similarly to most related studies, our solution uses a force sensor to detect push failures and empirically check the removability of blocks. On the other hand, our perception system presents several novelties: a block identification approach based on an instance segmentation neural network, an eye-in-hand camera configuration, and a visual servoing control for the manipulator.

\subsection{Deep Learning for Object Recognition}
\label{sec:related_instance}

During the last few years, deep learning \cite{lecun2015deep} has achieved state-of-the-art performance on various computer vision tasks. Different methods can be used to extract knowledge from visual data and give them a semantic interpretation. The literature refers to \textit{classification} as the task of assigning a descriptive label to the whole image. Several approaches have been proposed to solve this problem, introducing architectural methodologies as convolutions \cite{krizhevsky2017imagenet}, residual connections \cite{he2016deep}, feature \cite{hu2018squeeze} and space attention \cite{woo2018cbam}, or the more recent Transformer-based architecture with self-attention~\cite{vaswani2017attention, dosovitskiy2021image}.

Suppose a more fine-grained semantic description is needed. In that case, the \textit{object detection} task has the objective of localizing instances that belong to target classes with the regression of bounding boxes. This detection allows the system to understand the scene hierarchically depicted in the image, assign multiple labels, and spatially identify the objects in the image reference frame. Popular methodologies for object detection \cite{liu2016ssd, redmon2016you, redmon2017yolo9000, redmon2018yolov3} have focused on efficient and real-time execution to be used on continuous streams of images.

On the other hand, the \textit{semantic segmentation} task aims at assigning a semantic label at the pixel level by predicting masks that identify the portion of the image belonging to a certain class. Classical approaches to this task are based on fully convolutional networks organized in an encoder-decoder fashion~\cite{long2015fully, ronneberger2015u}, which adopt successive downsampling and upsampling operations to predict labels at the pixel level. 
The main difference between semantic segmentation and object detection is that the former does not identify single instances but only regions that depict objects of the target classes.

\textit{Instance segmentation} aims to localize single instances by predicting masks. This approach allows the most precise interpretation of the input image since it avoids coarse bounding box localization by identifying masks at a pixel level. Several methods have been proposed in the literature to solve this task. Mask-RCNN~\cite{he2017mask} extends an object detection method called Faster R-CNN \cite{ren2015faster} and is based on a two-stage approach that first proposes possible regions of interest (ROI) and predicts segmentation masks and classes in the second stage. Other approaches are based on one-stage architectures \cite{li2017fully, chen2018masklab} and are inherently faster than two-stage methods. Other approaches solve a semantic segmentation task and then perform instance discrimination with boundary detection \cite{kirillov2017instancecut}, clustering \cite{liang2017proposal} or embedding learning \cite{newell2017associative}. Recently proposed YOLACT \cite{bolya2019yolact} and YOLACT++ \cite{yolact-plus-tpami2020} focus on a real-time approach to instance segmentation that extends an object detection approach with mask proposals. The combination of mask proposals and bounding boxes gives pixel-level instance localization. In this work, we adopt this approach due to its computational efficiency and ability to detect many near objects, typical of Jenga block segmentation.

\subsection{Visual Servoing in Robotics}
\label{sec:related_visual_servoing}
Industrial robotic tasks such as assembly, welding, and painting represent the standard scenarios where manipulators execute repeatable point-to-point motion by using off-line trajectories \cite{evjemo2020trends}. However, the variability and disturbances of different environments may affect the estimation of the target pose and lead to a degradation of task accuracy.  Also, there are better strategies than this open-loop control technique for motion-based objects due to the target position and orientation variations.

Visual-based control strategies recently emerged as valid candidates for real-time trajectory computation and correction. 
In particular, visual servoing was introduced in 1979 \cite{VS} and refers to closed-loop systems where visual measurements are fed back into the controller to enhance task precision. Several works have proposed this approach for robotics applications in medical \cite{AzizianMahdi2014Vsim}, agricultural \cite{8605209}, and aerospace contexts. The ability to move a robotic arm flexibly in high-precision surgery operations \cite{StaubChristoph2010Aotp, VorosSandrine2010VRSH, KrupaA2003A3po} confirms the potential of this technique in complex scenarios where small errors can compromise human health.

Visual servoing taxonomy distinguishes two approaches \cite{HutchinsonS1996Atov} according to the type of tracker used to generate visual features. Image-based visual servoing reconstructs the relative pose of the target in the manipulator reference frame using the camera field of view. This approach is widely used in agriculture applications \cite{BARTH201671, MehtaS.S2016Rvsc}, where occlusions of the camera can lead to poor visual feature extraction. 

On the other hand, position-based visual servoing leverages a priori geometrical information on the object to derive the corresponding visual features. 
Recently, hybrid schemes have tried combining the two techniques and leveraging 2-D and 3-D visual features. In \cite{7361979}, an aerial vehicle with a robotic arm presents a hybrid visual servoing scheme to plug a bar into a fixed base. In this case, a marker detector provides the pose information of the object to be grasped, narrowing the possible field of use.  
Indeed, the robustness of tracking algorithms remains a central problem for visual-based control. Recently, in \cite{ComportA.I2006Rmtf}, the authors proposed a novel approach that uses Augmented Reality (AR) to generate either 3D model-based tracking or 3D model-free tracking techniques to enhance the system robustness. 

\section{Methodology}
\label{sec:methodology}

In this section, we frame the Jenga game and translate the rules of the game into a methodological set of requirements for the robotic system. First, a player's final goal is safely removing blocks from the tower. In the original game setting, a player has to place each extracted block at the top of the tower to validate its round and continue with a new block extraction. In our robotic experimental setting, we remove this rule and aim to extract as many blocks as possible from the tower without reallocation. This choice is mainly related to the limited workspace of the e.DO manipulator, designed for educational purposes, since it cannot reach the fallen blocks behind the tower.
Moreover, our custom end-effector, similar to a human finger, cannot perform grasping and instead extracts blocks by pushing. As the first practical task, a Jenga player should be able to select one of the blocks of the tower to be extracted. To this end, each block of the tower is identified in our system using an Instance Segmentation deep neural network. Moreover, we define a heuristic block selection policy based on the idea that extracting multiple blocks from the same tower level is not recommended. Moreover, as better detailed in \ref{sec:policy}, blocks at different tower layers present diverse frictions and effects on the tower's configuration.

Therefore, our solution is based on the following assumptions:
\begin{itemize}
    \item the identification and pose estimation of each block of the tower is the first step to selecting a suitable piece and approaching it;
    \item a removable block cannot be identified only by visual analysis: the integration of a tactile perception system is needed;
    \item a sufficiently precise alignment between the end-effector and the center of a target block allows the arm to extract it successfully by simply pushing.
\end{itemize}

A complete illustrative schematic of the system proposed to play Jenga is depicted in Figure \ref{fig:full-scheme}.

\subsection{Deep Instance Segmentation and Pose Estimation}
\label{sec:segmentation}

The proposed methodology's first step relies on identifying the blocks present in the tower by visual analysis. We adopt a deep learning-based method to perform instance segmentation. The necessity of using a deep learning approach is caused by the fact that wooden blocks do not have easily-distinguishable features while having predictable positions in the camera frame instead. In this setup, the vision system can benefit from the ability of neural networks to generalize to different light conditions and points of view. Moreover, training the model on an exhaustive dataset can make it robust towards missing blocks and tower misplacement during the game. Among different approaches for image semantic analysis, instance segmentation aims at detecting all the objects of interest in an image at the pixel level. Given an input image of the tower, the model should output a set of possible Jenga block candidates, together with their respective pixel masks. Instance segmentation can therefore be seen as a combination of an object detection task with a mask prediction task. By translating the position of the blocks from the image reference frame to the robot reference frame, it is possible to achieve block localization.

\paragraph{Model architecture}
We select the YOLACT++ \cite{yolact-plus-tpami2020} architecture to implement the instance segmentation algorithm. This architecture has been chosen for its computational speed, capability to handle occlusion, and efficiency in detecting a high number of tightly packed same-class objects. The model is based on a ResNet-50 \cite{he2016deep} + FPN \cite{lin2017feature} feature extraction backbone followed by two branches, one dealing with object detection and the other with mask prototype production. The detection branch outputs a set of anchor predictors $a$ as possible Jenga block candidates. Each anchor prediction consists of the class confidence $c$, four bounding box regressors, and $k$ mask coefficients. These mask coefficients are used to weigh the mask prototypes produced by the second branch. Both branches are based on convolutional layers applied to the features extracted by the common backbone.

The two branches are finally followed by a Mask Assembly block, which combines the masks with the coefficients predicted by each anchor to get to the final instances prediction. At inference time, anchor predictions are thresholded with a certain value $t_c$ on their confidence $c$ score to produce the actual output. Moreover, as in standard object detection algorithms \cite{liu2016ssd,redmon2016you}, an NMS (non-maximum suppression) method is applied to remove redundant predictions.

\begin{figure}[t]
    \centering
    \includegraphics[width=0.8\columnwidth]{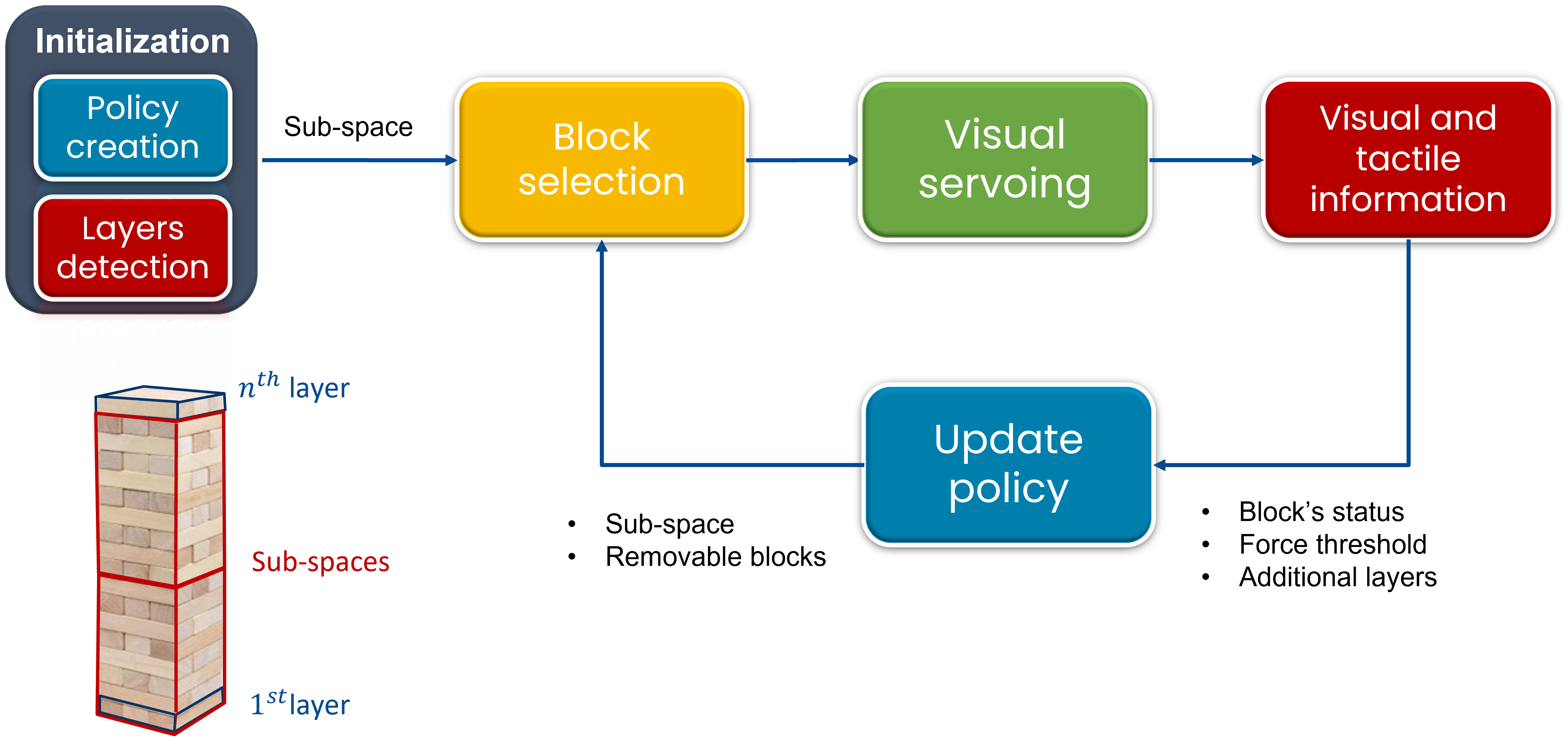}
    \caption{High-level input-output diagram describing the logic flow of the policy used to select the next block to extract from the Jenga tower. The layers of the tower are numbered from the $1^{st}$ to the $n^{th}$ starting from the bottom, as shown in the schematic on the left. The policy starts to extract the blocks from one of the two sub-spaces of the tower, randomly chosen. It proceeds until all levels contained in the sub-space get tested, to finally switch to the other sub-space trying to avoid the tower to collapse.}
    \label{fig:Policyscheme}
\end{figure}

\subsection{Block selection policy}
\label{sec:policy}
Our solution defines a heuristic block selection policy based on physical and empirical considerations. Visual information cannot provide sufficient information to determine the status of a block. The imperceptible tolerances of each block cause minuscule variations in pressure that prevent visual-based systems from understanding which blocks are truly removable. The blocks in the higher layers are easier to extract, i.e., they can be pushed out of the tower by applying a smaller force. However, pushing from a decentralized contact point contributes to the formation of torques on the block that causes asymmetry in the tower. This effect is amplified on upper layers due to lower friction forces and may affect the stability of either adjacent blocks or the tower itself. Instead, the friction force increases for blocks located at lower levels, making the extraction harder and risking the Jenga tower falling. 

Such considerations imply the need for a policy to select the block to extract. However, the policy must also consider the physical dimensions of our 6-DOF anthropomorphic manipulator. Indeed, during the extraction primitive, the robot has to be parallel to the block, which implies a loss of DOFs. These orientation constraints restrict the robot's workspace, so the manipulator can only reach a limited range of tower levels. In order to overcome these issues, the policy divides the entire tower into two sub-spaces according to the robot's workspace. In particular, the sub-spaces correspond to the upper and lower levels, each with a predetermined number of tower levels. By convention, the numeration of tower levels carries in ascending order, where one corresponds to the lowest level. In addition to this vertical division, the policy is also initialized with the block direction for each level. Indeed, two possible tower orientations have a relative rotation of $90\deg$ between the two. The pose estimation described in \ref{sec:pose_estimation}, applied by the convention on the top, provides the reference to infer the actual orientation of the tower and all its levels.

As human players usually do, we initially adopt a random policy that selects the Jenga pieces in one of the sub-spaces. Then, the manipulator approaches the chosen block and starts pushing. At this point, force sensor data is collected to evaluate the block's status according to the adopted force threshold, as explained in \ref{sec:force_sensor}. After each trial, a memory list updates the information on extracted and tested blocks. More in detail, a memory buffer stores the information about the blocks: the status (present, tested, or extracted), the threshold force to apply, and the current number of extraction attempts. In addition, it also keeps track of additional layers as the game goes on. Only one piece per level can be extracted as a further safety measure. 

The policy repeats the process until all levels contained in the sub-space get tested. After that, the system changes the sub-space to test additional levels until the tower collapses. More in detail, selecting the higher layers as the starting sub-space leads to three sub-spaces in total, with extra layers being included in the final sub-space. Alternatively, if the process begins by extracting blocks from the lower layers, there will be only two sub-spaces, with extra layers being included in in the high sub-space. Figure \ref{fig:Policyscheme} provides a minimal representation of the policy strategy. Except for the pick-and-place operation, the game implementation does not neglect any official rule.

\subsection{Block pose estimation}\label{sec:pose_estimation}
Predicted block masks are the input of a post-processing unit implemented using OpenCV. It estimates the position and orientation of the desired block to be extracted with respect to the camera. This information is the first requirement to approach the block's face starting the manipulator's tracking and visual servo control. The pose estimation is performed with an intrinsic calibrated camera through the Perspective n Points (PnP) algorithm optimized for planar points \cite{collins_bartoli_2014}. PnP is applied to the four corners of the front face of the target block, separately identified from the others thanks to the predicted segmentation mask, and paired with the known dimensions of the block. To increase the robustness of the tracking algorithm and the overall precision of the visual control of the manipulator, corners of nearby blocks are also considered (if present). Adjacent blocks may have diverse mask shapes: smaller for occluded front-facing blocks and larger for side-facing ones. Hence, two different sets of points are considered to estimate their poses. Side-facing blocks can provide a significant advantage in the subsequent tracking phase, offering a more stable visual reference during the motion of the end-effector.

Therefore, the PnP algorithm provides an initial estimate of the 6-DoF pose of the group of blocks (target and adjacent blocks) using segmentation mask corners to initialize the model-based tracker.

\begin{figure}[t]
    \centering
    \includegraphics[width=0.6\columnwidth]{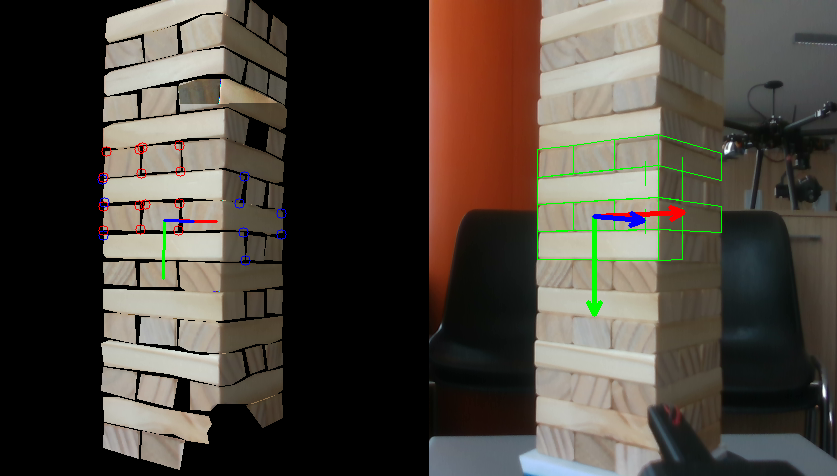}
    \caption{The initial pose estimation of the target block obtained by exploiting the segmented mask is shown on the left. The group model composed of multiple blocks to increase the robustness of the tracker is shown on the right.}
    \label{fig:tracking}
\end{figure}

\subsection{Tracking and visual servoing}
\label{sec:visual_servoing}
The geometric dimensions of a standard piece of Jenga $(25 \times 15 \times 75 $ mm) require precise movements to perform extraction successfully. Considering such dimensions and the width of the custom fingertip (i.e. 11 mm), it can be shown that the maximum position error from the center of the block must be smaller than 7 mm.
Therefore, the maneuver of the end-effector requires accurate trajectory planning to approach a block precisely. Standard point-to-point planning and online control \cite{2010R:mp} can be considered valid methodologies. However, these control schemes adopt an open-loop control system which requires high precision on pose estimation and tiny mechanical tolerances to reach the desired point with a small error. On the other hand, visual servoing is a closed-loop control strategy that exploits visual measurements to correct the pose of the end-effector with respect to the target in real time. Continuous visual feedback is used to correct the end-effector trajectory and to align the relative pose of the camera with the target block. For this reason, visual servoing is a competitive and flexible strategy to accurately approach the desired object (in this case, the Jenga block to extract).

In this regard, a robust tracking algorithm is a fundamental and challenging component of visual servoing control, which is required to guarantee smooth trajectories and allow a faster convergence to the target. As described in the previous sections, the segmentation masks are used to estimate the position and orientation of the desired block. At this stage, the tracking system receives both the initial pose estimation and the segmentation masks and combines them with a 3D model of a Jenga block.
Therefore, it detects the target to establish a continuous mapping of the 6-DOF pose of the Jenga block in the camera field of view. 
Specifically, the adopted stereo model-based tracking method VISP \cite{MarchandE2005Vfvs} combines several visual features, such as moving edges, key points, and depth information, to improve stability and robustness. The ViSP tracker requires the 3D model of a generic block to project its geometry into the image space and generate the visual features accordingly.

However, as mentioned in the previous Section \ref{sec:pose_estimation}, the tracking performance is only partially reliable if only the visible face of the desired block is used (Figure \ref{fig:tracking}). 
For this reason, the tracking algorithm does not rely only on the visual information of the single block to extract, but it integrates adjacent blocks in a unique group model. As already discussed in the previous subsection \ref{sec:pose_estimation}, this choice leads to more robust tracking thanks to the higher number of visual features extracted, especially from side-facing blocks. 

Moreover, the visual features of the group of blocks of interest are collected from different perspectives before the tracking is started. This acquisition phase enables to recovery tracking in case of sudden movements by exploiting the acquired pairings and re-initializing the block pose. 

Our approach adopts the eye-in-hand visual-servo configuration, whose extrinsic camera parameters are estimated based on the end-effector design. Thus, we indicate with $s = (x, y, \log(Z/Z^*), \theta_u)$ the visual features constructed for the $2$ $1/2$-D visual servoing tasks, where $(x, y)$ are the coordinates of the target in the camera reference frame, $Z$ and $Z^*$ are the current and desired depth of the point, respectively, and $\theta_u$ is a ${6x1}$ vector that identifies the rotation angles (expressed in the axis-angle convention) that the camera has to follow.  
Moreover, we refer to $\widehat{L_s^+}$ as the approximation of the interaction matrix and $e_s$ as the tracking error of the visual features, $s$, defined as $e_s = s-s^{\star}$, where $s^{\star}$ denotes the desired visual features.

The $6x1$ vector $v$ represents the linear and angular camera velocities that are computed through the following control law:
\begin{equation}
    v=-\lambda\widehat{L_s^+}e_s     
    \label{visualservoLaw}
\end{equation}
Hence, the above regulation control law defines the linear and angular velocities the camera has to follow to reach the target. Using forward kinematics, the vector $v\in \mathbb{R}^{6x1}$ is converted into the task space and executed through a motion rate controller using inverse differential kinematics. The overall closed-loop system runs at about 9 Hz on the laptop i7-8750H CPU. To further optimize the smoothness of the trajectory, we consider fixed constraints on the maximum joint velocities that the controller can predict. In Section \ref{sec:experiments}, we show that the visual servoing pipeline we adopted increases the overall accuracy and robustness of the Jenga extraction system.

\begin{figure}[t]
    \centering
    \includegraphics[width=0.6\columnwidth]{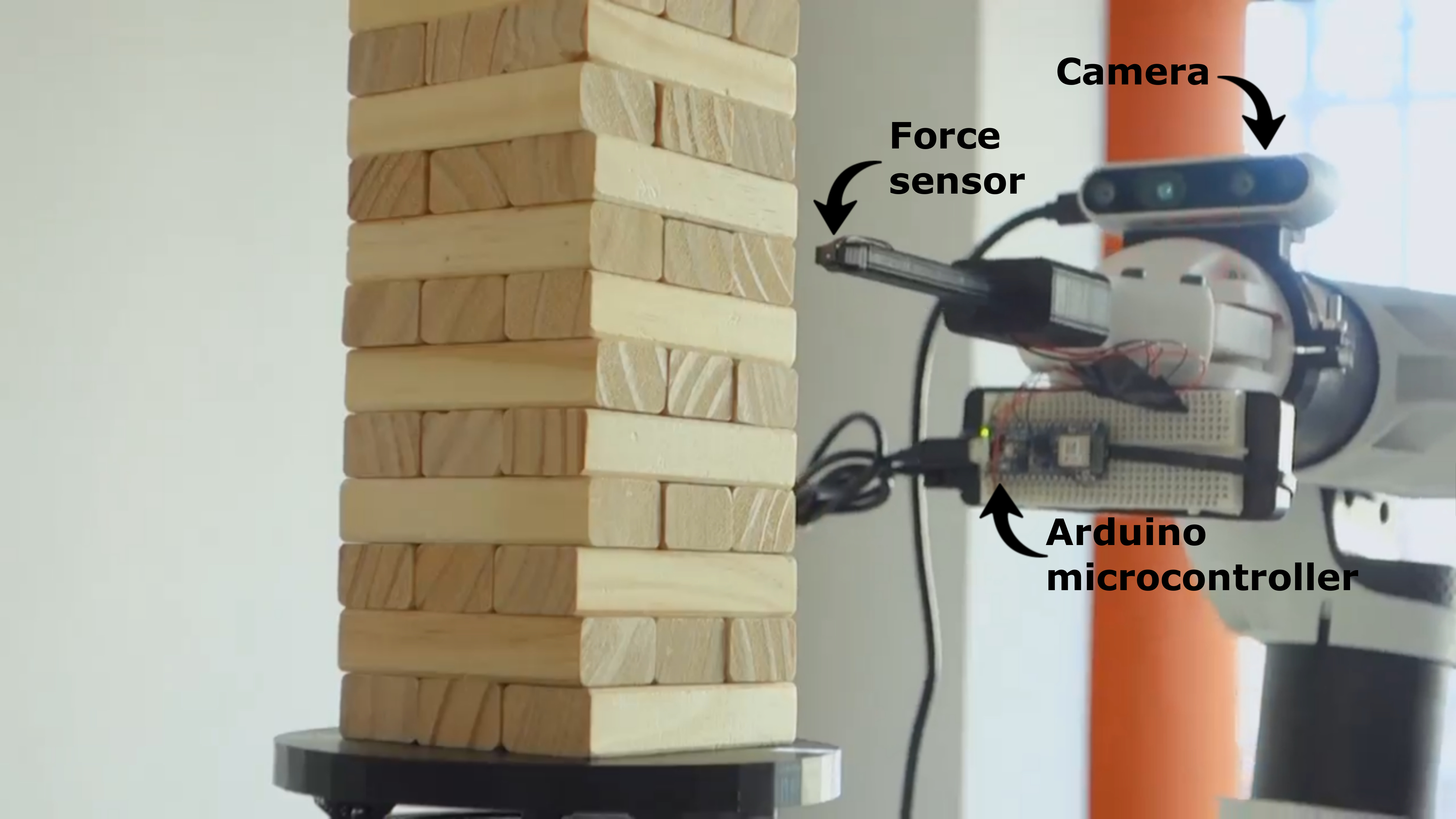}
    \caption{The manipulator has a human finger-inspired end-effector, which mounts a simple 1-D force sensor on its tip. The force sensor controlled by an Arduino Nano board reveals the status of a block (removable or not) once it is approached and pushed. An eye-in-hand camera configuration enables the visual control pipeline to align the fingertip with the Jenga piece accurately.}
    \label{fig:end_effector}
\end{figure}

\subsection{Tactile Perception}
\label{sec:force_sensor}
Tactile perception is a fundamental aspect of a human Jenga player since identifying potential removable blocks is not possible via visual analysis only. According to this, we decide to incorporate a low-cost 1-D force sensor in our system, which only provides information about the perpendicular reaction force between the end-effector and the target block. Hence, while previous works adopted a 6-D force sensor to adjust the direction of the end-effector while pushing \cite{mit_jenga}, we take advantage of our visual-based control strategy and guarantee a considerably good precision by simply pushing the block forward.
The one-dimensional force sensor is mounted directly on the fingertip of the end-effector and provides a digital output with a full-scale force span of $5 N$. The sensor is activated when an interaction between the fingertip and the block occurs. The block can be either stuck or free to move, so the push primitive uses the reaction force to derive a binary block classification. The closed loop explained in \ref{sec:visual_servoing} reads the measurements at 9 Hz while the Arduino microcontroller sends them at 20 Hz.

A challenging aspect of the game is that the more push attempts are performed, the more unstable the tower becomes, as extracted blocks or rotations perturb its structure due to pushes and retractions. Therefore, two thresholds detect immobile or moving blocks in $thr_{1}=0.32 N$ and $thr_{2}=0.18 N$. Such values are determined by combining theoretical \cite{force_based_jenga} and empirical results, better depicted in Section \ref{sec:experiments_force}. The threshold changes to a smaller conservative value in the second phase of the game when all the higher levels have been tested. Our choice can be defined as conservative, as it safeguards the stability of the tower rather than seeking more competitive performance.

\section{Experiments}
\label{sec:experiments}
The experimental setup includes the anthropomorphic educational manipulator, e.DO manufactured by Comau, a depth camera Intel RealSense D435i, a MicroForce FMA piezoresistive force sensor (5 N full scale, 12-bit resolution for 0.002 N sensitivity), an Arduino Nano 33 BLE, and 3D printed components such as a rotating base, an end-effector design extension, and camera support. The software runs on a single computer with an i7-8750H CPU and 16 GB of RAM. Thus, one of our goals is to investigate and prove the effectiveness of state-of-the-art with low-cost equipment. Visual control adopts an eye-in-hand configuration with camera support, while the small force sensor is placed on top of the end-effector extension connected to the Arduino Nano. 

\subsection{Reaction force threshold}
\label{sec:experiments_force}
In this section, we first describe the experiments carried out to define the static force threshold used to check the removability of blocks.
The real Jenga blocks present small differences in dimensions, generating a diverse pattern of friction forces in the tower each time it is rearranged for a new game. Although \cite{force_based_jenga} tried to provide a rigorous mathematical formulation of friction forces in the tower, we prefer an experimental approach to identify the correct threshold values to detect whether or not the robot push affects the stability of the tower.

\begin{figure}[t]
    \centering
    \includegraphics[width=0.7\columnwidth]{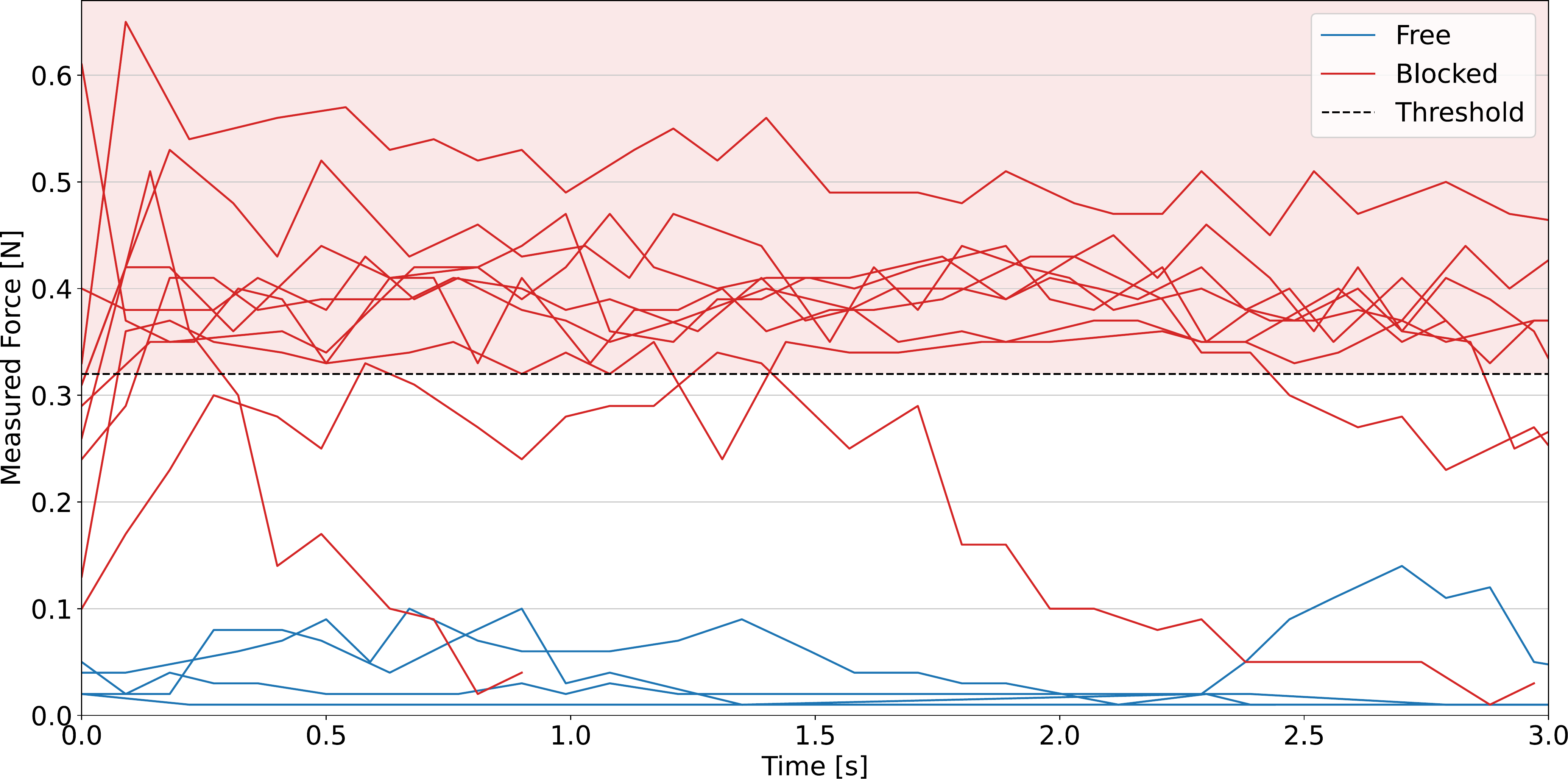}
    \caption{Reaction force measured over time during pushes for several blocks located in different layers and tower positions. The threshold force value of 0.32 N is also plotted to distinguish moving blocks in red from constrained ones in blue.}
    \label{fig:ForceThreshold}
\end{figure}

The measurements are taken with a complete tower configuration during the block extraction, starting the data collection of force reaction from when the contact between the block and end-effector starts and the force sensor detects a non-zero value. The experiment is run multiple times on different tower levels, mixing the blocks' disposition each time to test random friction conditions. 
The plot in Fig. \ref{fig:ForceThreshold} shows the force profile of 15 blocks located at different tower positions over time.

The choice of the threshold force values is not trivial, as it affects the system's ability to classify the block's removability state. By setting higher threshold values, more blocks can be classified as false positives, i.e., they are perceived as free to move, whereas they cannot be safely extracted. On the other hand, imposing lower threshold values affects system performance leading the system to avoid feasible block extractions.

As depicted in \ref{fig:ForceThreshold}, the initial threshold value is set to $0.32 N$, making the system more likely to detect false positives than false negatives. In addition, this is reflected in a more aggressive strategy at the beginning of the game, enhancing performance at the cost of tower stability. After attempting to remove half of the levels, the threshold is reduced to preserve tower stability. In fact, after removing a certain number of blocks, the tower becomes increasingly unstable, and the friction forces change according to the position of the extracted blocks. Due to the static friction force, the superposition of these effects increases the probability of disrupting the tower as soon as the contact between the fingerprint and block occurs. Therefore, mechanics and empirical observations led us to lower the threshold value to $0.18 N$. 
In general, the measurements are in the same range as the results obtained by the more extensive analysis of \cite{multiarticulated_jenga}, \cite{force_based_jenga}, and \cite{mit_jenga}, with forces between 0 N and 1 N and thresholds $\geq0.2$ N. 

\subsection{Instance segmentation experiments}
\label{sec:experiments_segmentation}

In this section, we report the method and the details of the procedure to train the instance segmentation model with a carefully devised synthetic dataset, as well as the results obtained by the experimental validation. The experimentation aims to assess the quality of the deep learning model and its generalization to the real pictures of our experimental environment.

\paragraph{Training setup}

The training of the instance segmentation model is entirely performed on a synthetic dataset crafted from a 3D model of the Jenga tower. Using the 3D modeling software Blender and its Python APIs through Blenderproc \cite{dlr139317}, hundreds of photorealistic and varied images of the tower are produced with automatic pixel-perfect annotations. This approach makes it fast and easy to obtain hundreds of samples without manually labeling real images. The training and validation datasets are composed of 800 and 80 images of size 640x480, respectively, rendered from a Blender synthetic scene composed of 48 cuboids arranged in a tower of 16 levels. We apply 12 different wood materials to the faces of these blocks to simulate the possible colors and wood line patterns with a realistic look. Each scene is loaded with blocks, a virtual camera, and a point light source. We design the following levels of scene randomization:
 
\begin{itemize}
    \item the materials are randomly sampled and assigned to all 48 cuboids to change their look;
    \item 6 to 24 cuboids are randomly displaced along their x and z axes by a distance between -4 mm and 4 mm;
    \item a point light source is positioned by randomly sampling a height of 0.1$\div$0.5 m spanning a circular arc of $60\deg$ centered on the tower with a random radius between 0.4 and 0.7 m;
    \item 2 to 9 random blocks are removed from the tower to create holes;
    \item camera position is sampled on a circular arc of $20\deg$ with a height between 0.05 and 0.2 m and a radius between 0.25 to 0.45 m.
\end{itemize}

The camera is placed to capture two faces of the tower at the same time. For each configuration, 10 samples are acquired with the full tower and 10 with random missing blocks.
The procedure is repeated for each random scene obtaining a diverse and complete dataset with different light conditions, block displacements, missing blocks, camera angles, and views of the experimental conditions. The scene's background is then filled with black, white, or gray.
The render time for all the 880 images was 4 hours on i7-9700K CPU.
The test set is composed of 20 real manually-annotated pictures from our experimental setup for a total of around 800 segmented tower blocks. The images are taken in slightly different light conditions and camera positions. 

The input 640x480 images are rescaled to 550x550 to be compatible with network input requirements. We adopt a ReseNet-50 backbone with pre-trained weights on ImageNet. We perform 8000 training iterations (69 epochs) with a batch size of 8, SGD optimizer with a momentum of 0.9, and a weight decay of $5\mathrm{e}{-4}$. The initial learning rate of $1\mathrm{e}{-3}$ is scaled down by a factor of 10 at iterations 5000, 6000, and 7000. We consider a positive intersection-over-union (IoU) value of 0.5 during training. The training is performed on a Tesla K80 GPU with Cuda 11.2.

\begin{figure}
\centering
\begin{subfigure}[t]{0.5\textwidth}
   \includegraphics[width=0.9\linewidth]{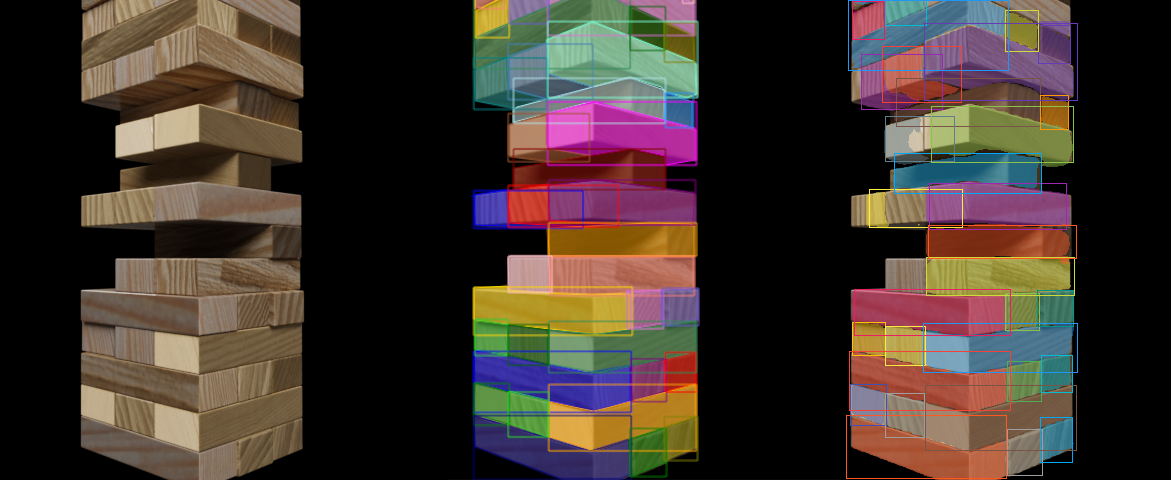}
   \caption{Synthetic Dataset (AP=69.9\% for 80\% IoU)}\label{fig:SegmentationValid}
\end{subfigure}%
\begin{subfigure}[t]{0.5\textwidth}
   \includegraphics[width=0.97\linewidth]{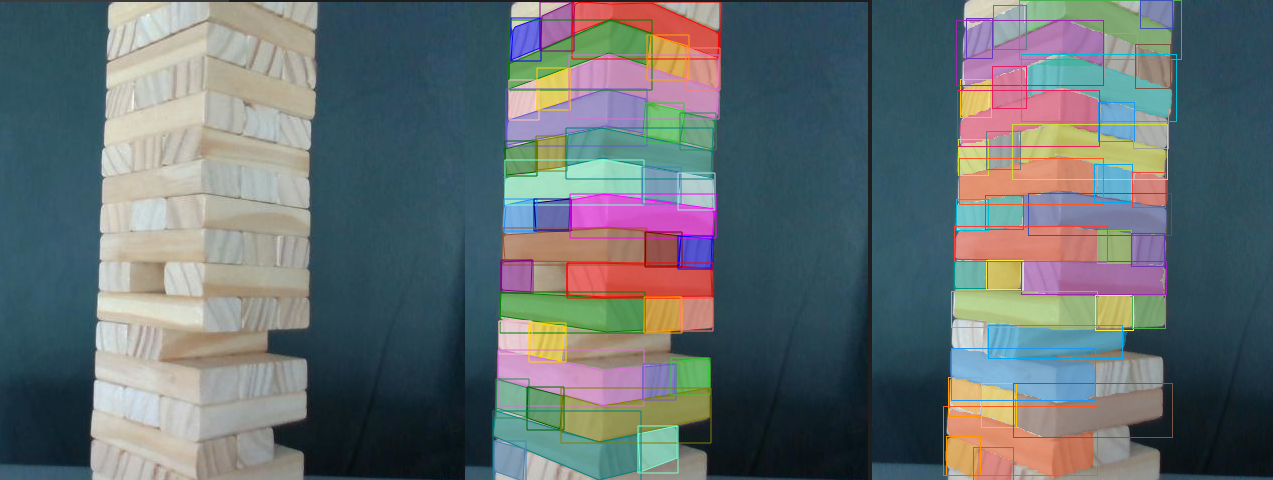}
   \caption{Real Images (AP=88.1\% for 80\% IoU)}\label{fig:SegmentationReal}
\end{subfigure}
\caption{Segmentation masks on simulated (a) and real (b) images. From left to right: source image, ground truth, and predicted masks.}\label{fig:Segmentation}
\end{figure}

\begin{table}[t]
\centering
\resizebox{0.4\columnwidth}{!}{%
\begin{tabular}{@{}lccc|c@{}}
\toprule
\textbf{Dataset} & \textbf{AP$_{\textbf{50}}$} & \textbf{AP$_{\textbf{80}}$} & \textbf{AP$_{\textbf{90}}$} & \textbf{Mean} \\ \midrule
Synth     & 90.08         & 87.76         & 63.2          & 78.37        \\
Real      & 75.98         & 53.40         & 11.53         & 53.09        \\ \bottomrule
\end{tabular}
}
\caption{Mask AP results at different values of IoU for the Jenga blocks instance segmentation model on both the synthetic and the real manually-annotated test datasets.}
\label{table:mAPcompare}
\end{table}

\paragraph{Instance Segmentation results}

As the main metric to assess the quality of the instance segmentation, we adopt the widely used Average Precision at different values of intersection-over-union (AP$_{\text{IoU}}$). A predicted mask is considered a true positive (TP) if it has an IoU with the ground-truth mask over the given threshold. AP is then computed as the area under the curve of the precision-recall plot obtained varying the confidence threshold $t_c$.

Tab. \ref{table:mAPcompare} reports the AP results at different IoU values, both on the synthetic test set and the real manually-annotated dataset. As expected, increasing the IoU threshold results in a performance drop due to the stricter requirements asked of the model. Generally, we observe a certain drop in performance when considering the real-world dataset, mainly due to border effects caused by approximate hand-made annotations and a decreased recall caused by the high number of instances in a single image. However, since a high recall is not required to perform block selection and tracking effectively, we state that the obtained real-world generalization is good enough for the target application. Visual comparison of a synthetic and a real image is reported in Fig. \ref{fig:SegmentationValid} and \ref{fig:SegmentationReal}.

\subsection{Tracking robustness}
\label{sec:experiments_tracking}
This experiment tests the robustness of the stereo model-based tracker in two different configurations to assess its ability to keep track of the object's pose during movements. In particular, we compare our experimental visual setup with the basic functionalities of the ViSP library \cite{MarchandE2005Vfvs}. The main difference lies in constructing the 3D block model and the tracker initialization method. A CAD model of the single block takes only into account the target piece and requires the user to initialize the model manually. Instead, our tracking system merges different pieces around the target block according to the tower arrangement described in \ref{sec:visual_servoing}. For this test, we exploit a rotating base to automatically turn the Jenga tower around its vertical axis by $45\deg$ clockwise or counterclockwise at a constant speed. This rotation brings one of the faces perpendicularly to the camera axis, thus keeping the target block always in the field of view.

While the tower rotates, spanning the whole angle range, we test the tracker to follow the block moving in the images. We measure the projection error $e_{proj}$ as the difference between the tracker's estimated rotation of the block and the actual rotation of the rotating base. Fixing the maximum acceptable error $err_{thr} = 25 \deg$, the failure condition is reached when $e_{proj} < e_{thr}$. For each run, we report the percentage of the total time (60 seconds) for which the tracker follows the block without failures, including the target block's tower level. Since this test's ultimate goal is to highlight the tracker's robustness, we keep the same threshold value used in the game. The trials are performed with two different angular velocities, $\omega_{1} = 2.5 \deg/s$ and $\omega_{2} = 8.3 \deg/s$ for the same target.

\begin{table}[t]
\centering
\resizebox{0.5\columnwidth}{!}{%
\begin{tabular}{@{}ccccc@{}}
\toprule
\textbf{Level} & \textbf{Vel {[}$\deg/s${]}} & \textbf{Single {[}\%{]}} & \textbf{Group {[}\%{]}}\\ \midrule
5        & $\omega_{1}$ & 36.9 & 100 \\
5        & $\omega_{2}$ & 17.0  & 100 \\
6        & $\omega_{1}$ & 13.1  & 100 \\
6        & $\omega_{2}$ & 8.0  & 100 \\
8        & $\omega_{1}$ & 28.9 & 100 \\
8        & $\omega_{2}$ & 10.6  & 100  \\
9        & $\omega_{1}$ & 19.4 & 100 \\
9        & $\omega_{2}$ & 29.5  & 100 \\
10       & $\omega_{1}$ & 18.4 & 100 \\
10       & $\omega_{2}$ & 8.1  & 100 \\
11       & $\omega_{1}$ & 24.7  & 100 \\
11       & $\omega_{2}$ & 14.8  & 100 \\ \bottomrule
\end{tabular}%
}
\caption{Largest tracking time comparison between the 3-D CAD models generated as a single block model and our pattern-based multi-block model.}
\label{tab:TrackingStability}
\end{table}

The results in Table \ref{tab:TrackingStability} suggest that ViSP \cite{MarchandE2005Vfvs} is a scalable library that can be further optimized according to the requirements of our task. Our tracking system not only overrides the point-to-point manual initialization leveraging the segmentation mask prediction, but it also significantly improves the tracking robustness up to 7.5 times.

\subsection{Visual servo convergence and accuracy}
\label{sec:experiments_visual_servo}

The visual servo control law is tested, in terms of time and spatial error, by bringing the end-effector to the computed goal. To this end, the tracking system estimates the block's pose (i.e., position and orientation) to extract the visual features and compute the $6x1$ velocity vector in the camera reference frame. Such velocities are then converted in the end-effector's reference frame through the extrinsic parameters, estimated on our custom fingertip's design. Then, the linear and angular velocities are translated into joint velocities and actuated accordingly.

\begin{figure}[t]%
\centering
\includegraphics[width=0.7\columnwidth]{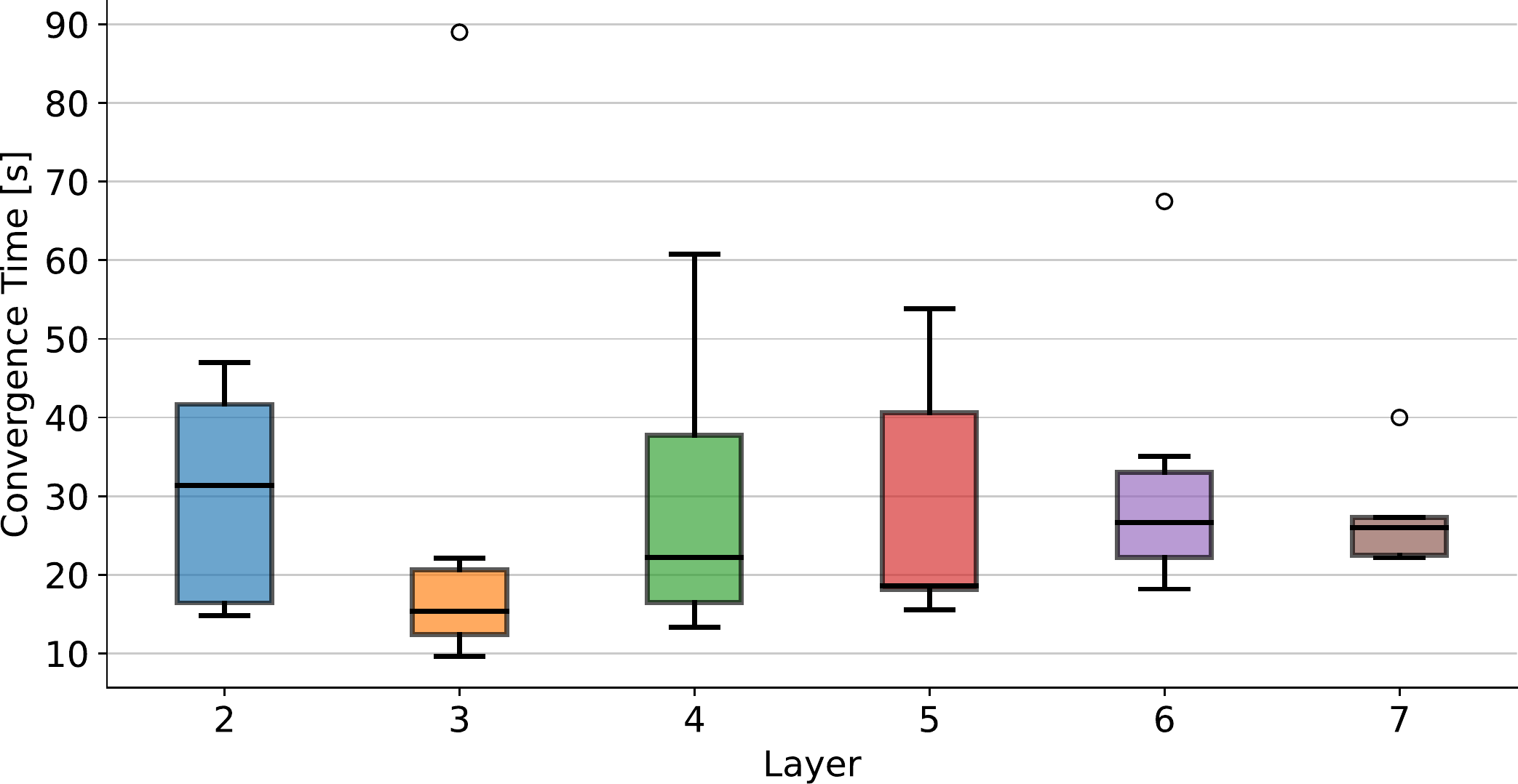}
\caption{Mean and standard deviation of convergence times for blocks on the same level, where the black line is the mean, and the dispersion of the values is shown.}\label{fig:VStimedeviation}
\end{figure}

\paragraph{Timing convergence}
The experiment begins with the robot in a default pose at the same height as the tower's first level. Then, a timer starts and automatically stops when the end-effector reaches the desired pose with a fixed tolerance on the visual servo error magnitude as $tolerance=0.00002$. This test is repeated for each tower level regardless of the block configuration. 
In Figure \ref{fig:VStimedeviation}, we report the distribution of convergence time at different tower levels. The average convergence time is roughly constant among levels, while we observe a high variance between different observations. This is caused by the fact that most of the convergence time is spent reaching the desired orientation rather than the desired position. The oscillations in the tracker estimate are higher when the camera is close to the tower, and the 3D block model degenerates to a plane face. The tracker is set to tolerate up to $25 \deg$ of mean reprojection error before failing, so the robot performs many corrections to the orientation, using all its joints to follow the oscillations. This sometimes generates longer convergence times. However, time is not a primary constraint in Jenga, so it is acceptable to trade convergence speed for more precise end-effector alignment.

\begin{figure}[t]%
\centering
\includegraphics[width=0.7\columnwidth]{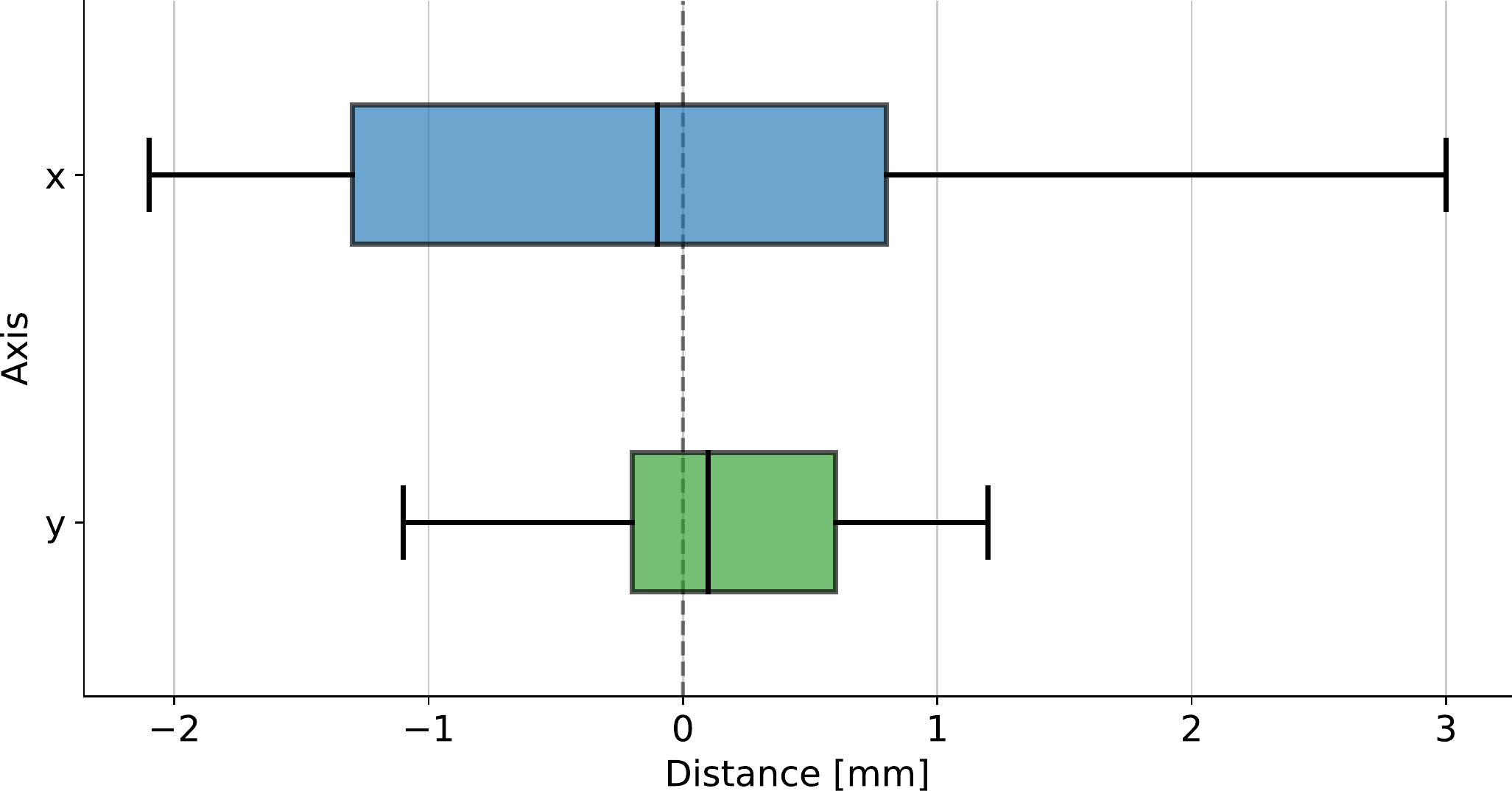}
\caption{Mean and standard deviation of the distance between the contact point of the end-effector and the center of the selected block.}\label{fig:VSAccuracy}
\end{figure}

\paragraph{Spatial accuracy}
The second part of the test aims to assess the manipulator's accuracy in pushing blocks. It consists in manually stopping the robot as soon as the end-effector reaches the target and measuring the distance between the contact point and the center of the block. Defining $err_x$  and $err_y$ as the horizontal and vertical errors, we perform a statistical analysis of the system precision. The results depicted in Fig. \ref{fig:VSAccuracy} are obtained from 21 trials on different blocks.
The results demonstrate the accuracy and repeatability of our system, as a mean error below 0.2 mm and a standard deviation below 1 mm are compatible with the accuracy defined in \ref{sec:visual_servoing}. Also, it is worth noting that the mean value for both axes is close to zero, which indicates the deep focus on estimating the extrinsic camera-to-robot and intrinsic pixel-to-mm parameters. The differences between the two axes mainly depend on the manipulator's dexterity and its mechanical tolerances.

\subsection{Consecutive block extractions}
\label{sec:experiments_extraction}

Finally, the different modules are integrated, and tests of the architecture are performed on the full system. The experiment includes 18 trials following the semi-automatic loop explained in Section \ref{sec:policy}. The goal is to extract as many blocks as possible without breaking the tower.

In order to properly validate our results, the chosen metric considers both correctly extracted blocks and correctly classified immobile blocks. Hence, an attempt is labeled as successful in one of the following cases:
\begin{enumerate}
    \item A mobile block is correctly pushed without perturbing the stability of the tower;
    \item The binary classification model correctly identifies the block as immobile, and the arm retracts without causing the tower to fall.
\end{enumerate}
After each attempt, the policy updates the block status obtained from visual and tactile measurements during the extraction process. In particular, force sensor data is used to detect the block status, while visual measurements provide feedback on the result of the extraction primitive. The latter integrates pose estimation information and the direct kinematics of the manipulator in order to verify the complete extraction of the block from the tower. The minimization of the angular tracking error in the visual servo law allows positioning the end-effector as parallel as possible to the block by pushing it out of the tower (\ref{visualservoLaw}) along the z-axis of the block. The robotic arm then retreats the long finger back along the same axis. This open-loop push primitive runs synchronously to the force sensor at a frequency of 20 Hz to eventually abort the execution if high reaction force values are detected. If the manipulator successfully extracts the block, the policy updates the memory buffer with the current block status and verifies the presence of other removable blocks in the sub-space in order to start a new attempt. Conversely, in the case of immobile blocks, the manipulator pursues to extract the remaining blocks of the same layer. In this scenario, the policy removes the block from the list of removable blocks for future extractions. 

In our setup, the operator has three interactions that make the game semi-autonomous: changing the tower's position to allow the robot to switch from higher to lower layers (or vice versa), placing the extracted blocks on top to form a new level, and aborting extractions when failures occur.

Minor failures do not stop the game, as they do not cause the tower to collapse. The operator can abort an attempt when the system fails for reasons unrelated to the extraction. In this case, the loop continues on the next iteration, and the policy is not updated.
Possible minor failures are:
\begin{itemize}
    \item The robot reaches a singularity position when the target block level is at the edge of the workspace;
    \item The tracking is lost as either the tower is in bad light conditions or the target block lies outside the field of view;
    \item The initial pose estimation is not precise enough because the segmentation model does not detect all the corners of the block correctly.
\end{itemize}

To minimize these failures and prevent the perturbation of the tower's stability, we make some conservative choices. Different force thresholds are assigned depending on the game's phase, i.e., which sub-space of blocks is currently tackled. This way, the policy addresses the first sub-space with an aggressive force threshold of 0.32 N. After all the first sub-space levels have been tested, the policy addresses the second sub-space with a cautious threshold of 0.18 N. The effect of adding the extracted blocks on top is twofold: their weight contributes to changing the friction between the blocks and shifting the center of gravity of the tower. Furthermore, each new level is considered in the policy for additional attempts. This way, we obtain a minimum of 42 attempts (3 per level) and 14 block extractions (1 per level), plus 3 additional attempts and 1 extraction for every newly formed level.

\begin{figure*}[t]
\centering
\begin{subfigure}{0.5\columnwidth}
  \centering
  $\vcenter{\hbox{\includegraphics[width=0.6\columnwidth]{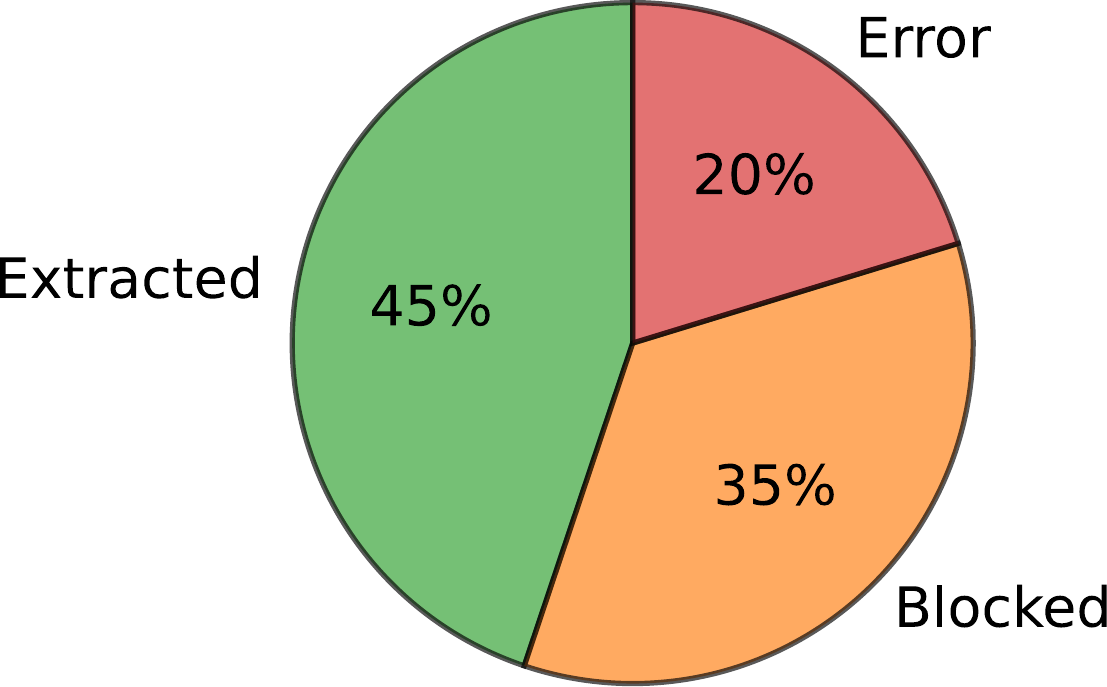}}}$
  \label{fig:sub1}
\end{subfigure}%
\begin{subfigure}{0.5\columnwidth}
  \centering
  $\vcenter{\hbox{\includegraphics[width=0.9\columnwidth]{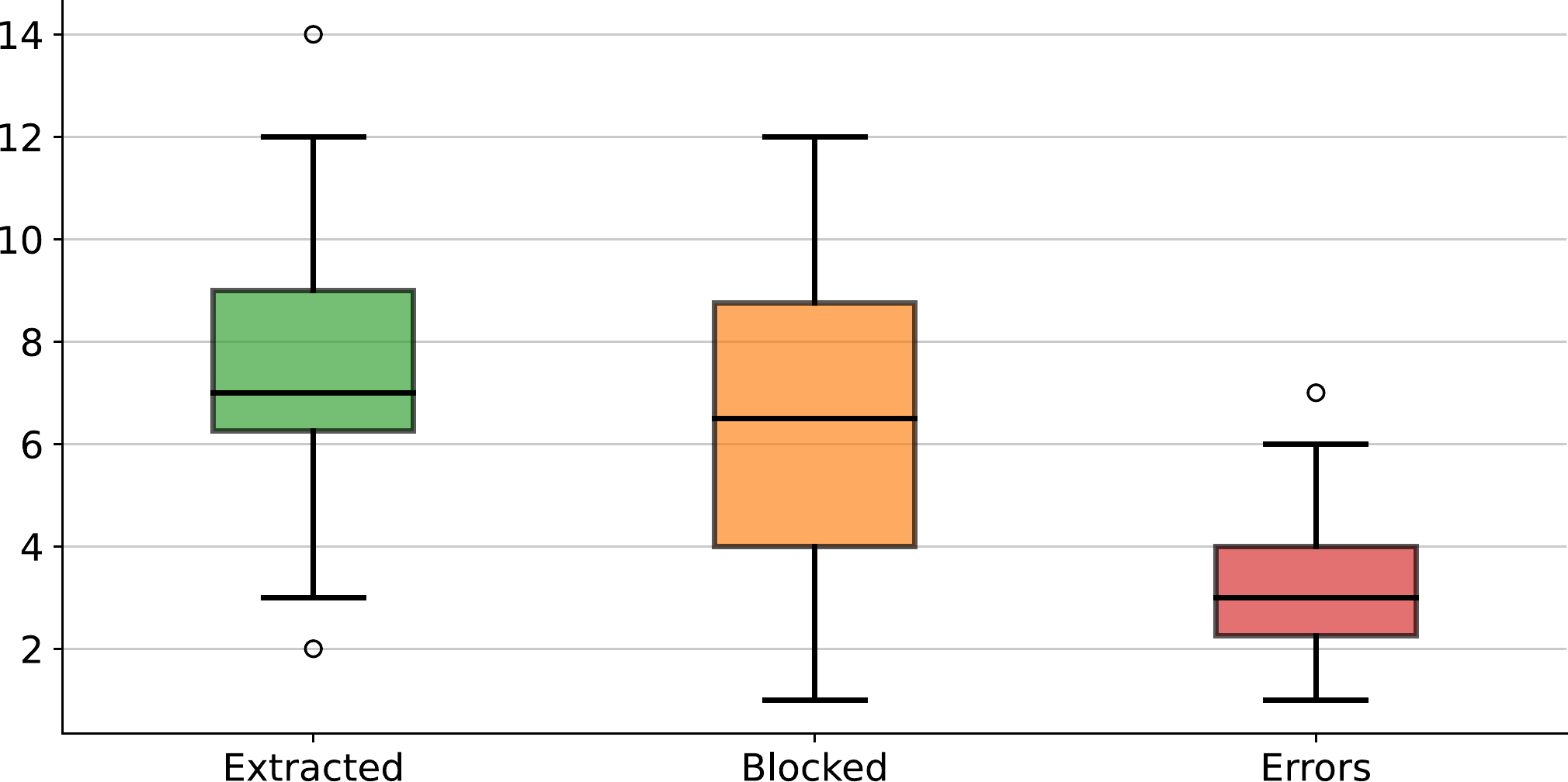}}}$
  \label{fig:sub2}
\end{subfigure}
\caption{The plot on the left illustrates the average number of extracted blocks(green), the blocks correctly classified as unmovable(orange), and errors among 18 trials (red). On the right, the statistical distribution of the outcomes is plotted. Our system can perform successful extraction or correctly identify unmovable blocks in 80\% of the attempts, achieving the highest score of 14 extracted blocks.}
\label{fig:final_test}
\end{figure*}

Results are reported in Figure \ref{fig:final_test} with details on the distribution of complete extractions, unmovable blocks correctly classified, and attempts ended with an error.
Experiments show that 80\% of the extraction attempts are successful. More specifically, 45\% of the attempts lead to extracting the block correctly, while 35\% of them find an unmovable block.
The errors (20\%) are mainly caused by the system losing the tracked block or by an imprecise initial pose estimation. Each time an error occurs, the current extraction attempt is aborted, and a new one is started.

Each experiment ends when the tower falls. In our experiments, the fall occurred after 13.8 correct attempts on average (7.5 correct extractions and 6.3 detected unmovable blocks, respectively). In most experiments, the fall was caused by the increasing instability of the tower after multiple extractions. However, a few experiments ended earlier because of poor tracker performance in providing the position and orientation during the push movement or incorrect detection of mobile blocks (two experiments ended after only 5 extraction attempts).
The highest registered score counts 14 successful extractions and 11 detected unmovable blocks.
The results are comparable to those of \cite{force_based_jenga}, which scored 14 consecutive extractions using a 6-axis force sensor while being much higher than the 5 extractions of \cite{strategic_jenga}, which also had inexpensive equipment. The score is far behind both the 20 and the 29 of respectively \cite{mit_jenga} - more extensive architecture and artificial intelligence - and \cite{manipulator_plays_jenga} - more complex hardware and simpler Jenga game setup.

\section{Conclusion}
\label{sec:conclusion}
In this paper, we propose a complete system to play the challenging game of Jenga with cost-efficient robotic hardware. The contribution of this work goes beyond the Jenga game, proposing an advanced, adaptable solution for accurate manipulator control in delicate robotic tasks. We demonstrate that a visual-based approach for perception and control can provide the robot with significant benefits in terms of scene understanding and control accuracy.

Differently from previous works, the main components of our system are a Deep Instance Segmentation neural network used to identify each block of the Jenga tower and a visual tracking and control pipeline to continuously adjust the pose of the end-effector as it approaches the block. A low-cost 1-D force sensor is integrated into the system to check the removability of the target block, drastically reducing the cost and complexity of the overall system. Our extensive experimentation shows the remarkable performance of each fundamental component of the solution. After examining the accuracy and stability of the perception and control units, we evaluate the whole system by playing Jenga and reaching a maximum of 14 successive block extractions. 
Future works may see the advancement of the reasoning capability of the robot: reinforcement learning agents can be investigated to optimize the planning policy for the game. Moreover, a visual-based sensorimotor agent could eventually replace the controller by directly mapping segmented and depth images to velocity commands for the end-effector.


\section*{Acknowledgements}
This work has been developed with the contribution of the Politecnico di Torino Interdepartmental Centre for Service Robotics (PIC4SeR)\footnote{https://pic4ser.polito.it/} and SmartData@Polito\footnote{https://smartdata.polito.it/}.

\bibliographystyle{abbrvnat}  
\bibliography{bibliography.bib}

\end{document}